# Respond-CAM: Analyzing Deep Models for 3D Imaging Data by Visualizations

Guannan Zhao[1*], Bo Zhou[2*], Kaiwen Wang[2], Rui Jiang[1], and Min Xu[2(✉)]

[1] Department of Automation, Tsinghua University, Beijing, China
[2] School of Computer Science, Carnegie Mellon University, Pittsburgh, PA, USA

**Abstract.** The convolutional neural network (CNN) has become a powerful tool for various biomedical image analysis tasks, but there is a lack of visual explanation for the machinery of CNNs. In this paper, we present a novel algorithm, Respond-weighted Class Activation Mapping (Respond-CAM), for making CNN-based models interpretable by visualizing input regions that are important for predictions, especially for biomedical 3D imaging data inputs. Our method uses the gradients of any target concept (e.g. the score of target class) that flows into a convolutional layer. The weighted feature maps are combined to produce a heatmap that highlights the important regions in the image for predicting the target concept. We prove a preferable sum-to-score property of the Respond-CAM and verify its significant improvement on 3D images from the current state-of-the-art approach. Our tests on Cellular Electron Cryo-Tomography 3D images show that Respond-CAM achieves superior performance on visualizing the CNNs with 3D biomedical images inputs, and is able to get reasonably good results on visualizing the CNNs with natural image inputs. The Respond-CAM is an efficient and reliable approach for visualizing the CNN machinery, and is applicable to a wide variety of CNN model families and image analysis tasks. Our code is available at: https://github.com/xulabs/projects/tree/master/respond_cam

## 1 Introduction

3D imaging data is commonly used in biomedical research. Since biological structures are 3D in nature, 3D images capture substantially more information as compared to 2D images. Recently, convolutional neural network (CNN) has become one of the most powerful tools for analyzing 3D imaging data in biomedical research, especially for tasks like classification, object detection and segmentation [4]. However, there is lack of explanation for what exactly CNNs have learned, how they learn, and how to improve them. To make CNN models more interpretable and reliable, the visualization of deep models is desirable.

A number of methods have been proposed to visualize CNNs for understanding computer vision tasks in 2D natural images. Most of the state-of-the-art works have been well summarized and integrated in [5]. Zeiler & Fergus proposed a method that mapped activations back to the input pixel space via Deconv-net, so that certain class-discriminative patterns in the input image were highlighted and visualized [8]. Class Activation Mapping (CAM) was proposed to highlight discriminative image regions for classification tasks [10]. However, it was only applicable to CNNs without fully-connected layers. Selvaraju [6] generalized the CAM and proposed the Grad-CAM, which provided informative visualization for analyzing and diagnosing deep models for natural image classification and visual question answering. As to our knowledge, the visualization of CNN in biomedical applications, especially regarding 3D imaging data, has not been properly studied.

---

[*] contributed equally

In this paper, we present a novel CNN visualization algorithm called Respond-CAM. The Respond-CAM is easy to implement, compatible with any CNN and is especially suited for CNNs using 3D biomedical images. We tested our Respond-CAM on the 3D images of macromolecular complex structures obtained from Cellular Electron Cryo-Tomography (CECT), an emerging imaging technique for visualizing these structures at their near-native state and at nanometer resolution. To show that our Respond-CAM is effective for a variety of inputs to the CNNs, we chose to use these macromolecular complexes for their large variations in shapes and sizes. This makes the CECT dataset ideal for testing the performance of the Respond-CAM. Previous studies have shown that CNN has important applications in CECT, including cellular structure segmentation, classification, etc. [7,2] We applied our visualization methods to two important CECT CNN models, denoted CNN-1 (taken from [7]) and CNN-2. We demonstrate that Respond-CAM achieves better visualization results than Grad-CAM, the current state-of-the-art method. Our theoretical analysis and experimental results both show that Respond-CAM has a preferable *sum-to-score* property, which makes the visualization results more consistent with CNN outputs than Grad-CAM [6]. These results indicate a significant improvement from Grad-CAM, especially in 3D biomedical images. In summary, our contributions are listed as follows:

- We propose a novel visualization approach (Respond-CAM) to analyze CNN models for 3D biomedical images, such as CECT;
- We prove the *sum-to-score* property of Respond-CAM and verify its significant improvement compared with current state-of-the-art approach;

## 2 Methods

Inspired by the work of [6], we propose the Respond-weighted Class Activation Mapping (Respond-CAM) to visualize and highlight the class-discriminative parts in a 3D image. The Respond-CAM can be easily extended to other tasks, such as segmentation and regression. Figure 1 gives an overview of our Respond-CAM architecture and its relationship to CNNs in various subtomogram analysis tasks. A *subtomogram* is a cubic subvolume of a CECT tomogram that is likely to contain only one macromolecule.

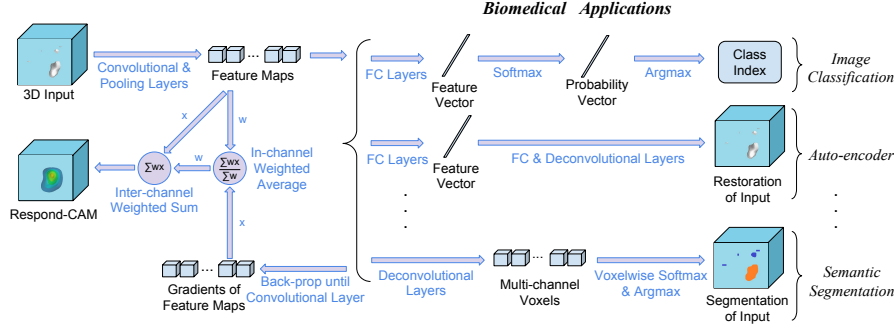

**Fig. 1.** Overview: for a trained CNN, we choose a differentiable scalar output of interest $y$ (e.g. an entry of the feature vector) and calculate the gradient of $y$ to the feature maps. Then we combine the feature maps and their gradients to get the Respond-CAM representing which areas in the input contribute the most to the output $y$.

For a class index $c$ and an output $A$ of a certain layer, we denote the corresponding Respond-CAM as $L_A^{(c)}$. Typically, we set $A$ to be the last convolutional layer [6].

We first compute the gradients of $y^{(c)}$ (the *score* for class $c$ before the softmax) with respect to all feature maps of the convolutional layer $A$. The gradients are denoted as $\frac{\partial y^{(c)}}{\partial A_{i,j,k}^{(l)}}$, where $A_{i,j,k}^{(l)}$ stands for the position $(i,j,k)$ of the $l$-th feature map. For every feature map $A^{(l)}$, the forward activations are fairly smooth because of the spatial correlation of convolution operations. However, their gradients, fed backwards through fully-connected layers, can be quite noisy and scattered. In order to "smoothen" the gradients, we calculate the $\alpha_l^{(c)}$ which is the weighted-average of all the gradients in the feature map using equation 1:

$$\alpha_l^{(c)} = \frac{\sum_{i,j,k} A_{i,j,k}^{(l)} \frac{\partial y^{(c)}}{\partial A_{i,j,k}^{(l)}}}{\sum_{i,j,k} A_{i,j,k}^{(l)} + \epsilon} \qquad (1)$$

where $\epsilon$ is a sufficiently small positive number for numerical stability and can be usually ignored. This $\alpha_l^{(c)}$ serves as an estimation of the "importance" of feature map $A^{(l)}$. The Respond-CAM $L_A^{(c)}$ has the same shape as every $A^{(l)}$, where $l$ is the index of the feature map. We then take a linear combination of $A^{(l)}$ to compute the Respond-CAM:

$$L_A^{(c)} = \sum_l \alpha_l^{(c)} A^{(l)} \qquad (2)$$

Although the result of Respond-CAM $L_A^{(c)}$ typically has a smaller shape than the input image does, we can resize it by interpolation in order to overlap it on the input image as a heatmap, as shown in the rightmost part of Figure 2.

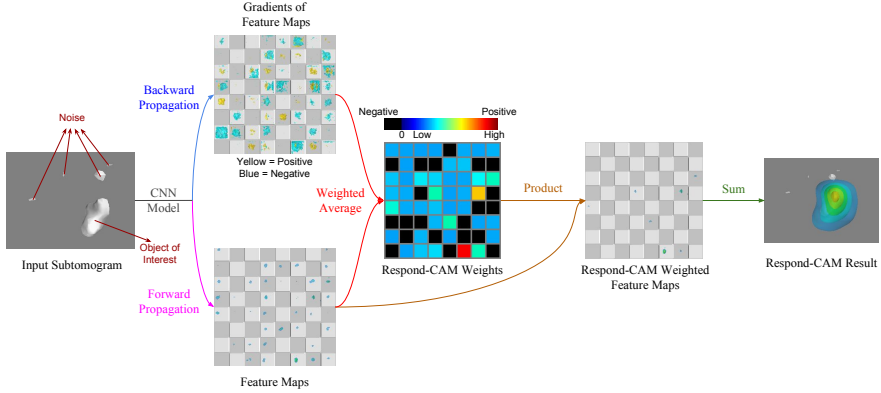

**Fig. 2.** Respond-CAM walk-through with an example. By applying the weighted average with each feature maps to their corresponding gradients, we get the Respond-CAM weights for each feature map. We calculate the Respond-CAM using equation 2 where the map with the largest weight (marked with red) contributes the most. As a result, the object-of-interest area gets highlighted.

The calculation process of Respond-CAM is demonstrated in Figure 2. Our approach is named "Respond-CAM" because each of the weights $\alpha_l^{(c)}$ can be interpreted as the *responsiveness* of its corresponding feature map $A^{(l)}$: it shows how rapidly, and in which direction, the score of class $c$ would change (i.e. *respond*) as the feature map $A^{(l)}$ changes.

We summarize three main advantages of Respond-CAM. 1) The algorithm is adaptive to any CNN architecture. 2) The calculation is efficient and simple to implement. 3) It approximately meets the *sum-of-scores* property better than the Grad-CAM.

**The sum-to-score property of Respond-CAM**. To demonstrate this property, we take CNNs with only one fully-connected layer as example. In such case, the gradients $\frac{\partial y^{(c)}}{\partial A^{(l)}_{i,j,k}}$ are exactly the weights of that fully-connected layer, which are constant once the CNN is trained. Therefore, it can be proved that the class score $y^{(c)}$ equals to the sum of Respond-CAM plus bias $b^{(c)}$ for the class (see Supplementary Section 1 of [9] for details), as shown in the following equation:

$$y^{(c)} = b^{(c)} + \sum_{i,j,k} (L_A^{(c)})_{i,j,k} \approx \sum_{i,j,k} (L_A^{(c)})_{i,j,k} \quad (3)$$

The equation indicates that the Respond-CAM approximately sum to the class score. This is denoted as the *sum-to-score* property. Then, we examine CNNs which contain multiple fully-connected layers. Although the gradients $\frac{\partial y^{(c)}}{\partial A^{(l)}_{i,j,k}}$ are not constant for such CNNs, we find the approximation still reasonable. The detailed discussions about the approximation shown above can also be found in Supplementary Section 1 of [9].

## 3 Experiments and Results

### 3.1 Data preparations

We simulated subtomograms in a similar way as previous works [1]. Each subtomogram consisted of $40^3$ voxels. We set the tilt angle to $\pm 60°$, and the signal-to-noise ratio to $+\infty$ and 0.1 respectively. Two datasets, including *the noise-free* and *the noised* data, were collected. In each dataset, we acquired 23000 subtomograms of 23 structural classes. We separated the subtomograms into the training set and the test set with a ratio of 4:1. We train our CNN models using 80% subtomograms of the training set for fitting and the rest for validation. Only the data from test sets are used for visualizations in our experiments.

We adopt two CNN models for our experiments. One of them is the same model that achieved the best classification accuracy in [7] with simple architecture, which we denote as *CNN-1*. Another is a slightly modified version of CNN-1. It further increases the classification accuracy on our datasets to over 96% and is denoted as *CNN-2*. Their architectures are described in detail in Supplementary Section 2 of [9].

### 3.2 Experimental results

Three representative examples that stress the difference between our Respond-CAM and Grad-CAM are shown in Figure 3. The last convolution layer is used here. The subtomograms are classified correctly by both CNN-1 and CNN-2, so it is expected that the heatmaps highlight the object of interest with maximum positive values. However, as shown in Figure 3, Grad-CAM sometimes gives noisy or even opposite results for the CECT volumetric data when noise is introduced, especially when applied to CNN-1, where the objects of interest are highlighted with negative values. In comparison, Respond-CAM produces significantly better heatmaps that indicate the locations of objects of interest with clean visualization result. One example of the visualization results for CNN-1 intermediate layers are also shown in Figure 4.

The Respond-CAM and the Grad-CAM were also evaluated quantitatively by examining their *sum-to-score* property. Intuitively, the better the *sum-to-score* property is approximated, the more the visualization results would be consistent with the CNN outputs themselves. The L1 error and Kendall's Tau (KT) [3] were calculated over both trained models using all subtomograms in the test set. These two metrics were chosen to verify the *sum-to-score* property from different aspects. L1 error measures the *absolute difference* between the class score and the sum of Respond-CAM/Grad-CAM heatmap, while the KT checks their *relative consistency*, i.e. we expect that the top-$n$ predictions are consistent whether they are sorted by the class scores or by the sums of heatmaps. For each subtomogram, the L1 error is defined using equation 4:

$$L_1 = \frac{1}{C} \sum_c \left| \sum_{i,j,k} L_A^{(c)} - y^{(c)} \right| \qquad (4)$$

where $\sum_{i,j,k} L_A^{(c)}$ is the sum of Grad-CAM/Respond-CAM for class $c$, $y^{(c)}$ is the corresponding score of class $c$, and $C$ is the number of classes. The $L_1$ is ranged in $[0, +\infty]$ and reaches 0 for a perfect consistence.

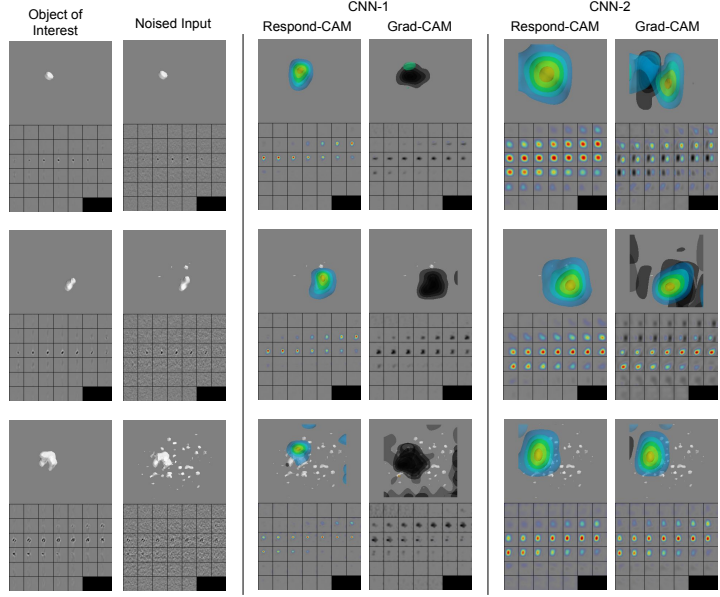

**Fig. 3.** Three cases in which Respond-CAM displays significant improvement. In each sub-figure, the same 3D image is presented in two ways. 1) Upper half: the parallel projection of subtomogram isosurface or Respond-CAM/Grad-CAM contours; 2) lower half: the corresponding 2D slices. Notice that heatmaps are resized so they can be overlaid on subtomograms.

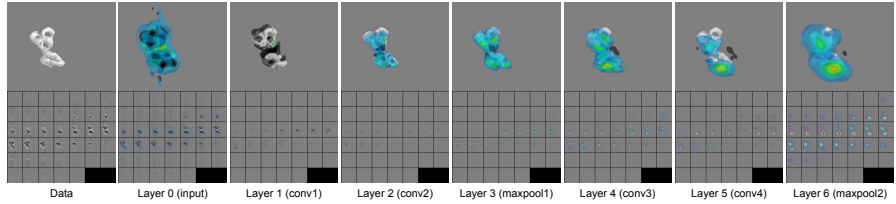

**Fig. 4.** The intermediate layer visualization of Respond-CAM using same classification network shown in Figure 3. All layers before fully-connected layer are visualized.

The KT measures the similarity of two sequences of class indexes sorted by a) scores and b) the sum of Grad-CAM/Respond-CAM. It is defined in equation 5:

$$\tau = \frac{P - Q}{P + Q} \quad (5)$$

where $P$ is the number of concordant pairs, and $Q$ is the number of discordant pairs. The $\tau$ is ranged in $[-1, 1]$ and reaches 1 for a perfect consistence. The evaluation results are listed in Table 1. We can see that the L1 error of our Respond-CAM is significantly smaller than the L1 of Grad-CAM. The KT of our Respond-CAM is also much closer to 1 as compare to results from Grad-CAM. Thus, the Respond-CAM much more strongly exhibits the *sum-to-score* property than Grad-CAM on both CNN models in CECT data.

**Table 1.** Comparison between Grad-CAM and Respond-CAM on L1 error and Kendall's Tau (KT) on different CNNs and datasets.

| Model | CNN-1 | CNN-2 | CNN-1 | CNN-2 |
|---|---|---|---|---|
| SNR | $+\infty$ | $+\infty$ | 0.1 | 0.1 |
| L1, Grad | $15.344 \pm 7.789$ | $8.117 \pm 3.854$ | $21.142 \pm 8.511$ | $14.020 \pm 5.232$ |
| L1, Respond | $\mathbf{0.601} \pm 0.187$ | $\mathbf{0.770} \pm 0.366$ | $\mathbf{1.006} \pm 0.721$ | $\mathbf{0.542} \pm 0.194$ |
| KT, Grad | $0.273 \pm 0.350$ | $0.822 \pm 0.097$ | $-0.116 \pm 0.341$ | $0.458 \pm 0.260$ |
| KT, Respond | $\mathbf{0.981} \pm 0.014$ | $\mathbf{0.976} \pm 0.018$ | $\mathbf{0.975} \pm 0.024$ | $\mathbf{0.984} \pm 0.013$ |

## 4 Discussion and Conclusion

We presented a novel CNN visualization approach that can be applied to CNN models with 3D/2D image inputs. Specifically, we used CECT data as the example to test our approach. Both visualization examples and quantitative tests show that our Respond-CAM achieves significantly better CNN visualization results as compared to the current state-of-the-art approach (Grad-CAM) on the 3D imaging data. In the Grad-CAM, the weights are calculated by simply averaging the gradients of the feature maps, while ignoring the feature maps themselves. This information loss in Grad-CAM sometimes produces improper weights and leads to undesirable results, such as those shown in Figure 3. Our experimental results also show that the Respond-CAM satisfies the *sum-to-score* property, which enables a more consistent and robust visualization of CNNs than Grad-CAM. Therefore, the Respond-CAM is a more reasonable visualization approach for CNNs, especially in the cases with 3D biomedical images, such as CECT. More details are available on [9].

In conclusion, our Respond-CAM achieved superior performance on visualizing the CNNs with biomedical 3D imaging data inputs. It is able to produce reasonably good results on visualizing the CNNs with 3D imaging data and can be applied to wide-variety of other imaging data. We believe the Respond-CAM is an efficient and reliable approach for visualizing the CNN machineries.

## 5 Acknowledgements

We thank Dr. Xiaodan Liang for suggestions. This work was supported in part by U.S. National Institutes of Health grant P41 GM103712. Min Xu acknowledges support from Samuel and Emma Winters Foundation. Rui Jiang is a RONG professor at the Institute for Data Science, Tsinghua University.

# Respond-CAM: Analyzing Deep Models for 3D Imaging Data by Visualizations – Supplemental Document

## 1 The Proof of the Sum-to-score Property of Respond-CAM

We give a detailed proof of the sum-to-score property of Respond-CAM, as well as some assumptions and discussions.

For CNN models that contains only one fully-connected layer, the class score is a linear combination of every voxel in every feature map in the last convolutional layer:

$$y^{(c)} = b^{(c)} + \sum_l \sum_{i,j,k} A^{(l)}_{i,j,k} (w^{(l)}_{i,j,k})^{(c)} \qquad (1)$$

where $b^{(c)}$ and $(w^{(l)}_{i,j,k})^{(c)}$ are the parameters of the last layer in the CNN model. Thus we have:

$$\frac{\partial y^{(c)}}{\partial A^{(l)}_{i,j,k}} = (w^{(l)}_{i,j,k})^{(c)} \qquad (2)$$

which is *constant* when the model has been trained.

Meanwhile, Respond-CAM is calculated as :

$$\alpha^{(c)}_l = \frac{\sum_{i,j,k} A^{(l)}_{i,j,k} \frac{\partial y^{(c)}}{\partial A^{(l)}_{i,j,k}}}{\sum_{i,j,k} A^{(l)}_{i,j,k} + \epsilon} \qquad (3)$$

$$(L^{(c)}_A)_{i,j,k} = \sum_l \alpha^{(c)}_l A^{(l)}_{i,j,k} \qquad (4)$$

where $\epsilon$ is a sufficiently small number that can be ignored.

Thus we leave out $\epsilon$, and let:

$$\beta^{(c)}_l = \sum_{i,j,k} A^{(l)}_{i,j,k} \frac{\partial y^{(c)}}{\partial A^{(l)}_{i,j,k}} \qquad (5)$$

Using Equation 2 and 5, Equation 1 can be now rewritten as:

$$y^{(c)} = b^{(c)} + \sum_l \beta^{(c)}_l \qquad (6)$$

while Equation 3 can be rewritten as:

$$\beta^{(c)}_l = \alpha^{(c)}_l \sum_{i,j,k} A^{(l)}_{i,j,k} \qquad (7)$$

Using Equation 7, we rewrite Equation 6 as:

$$y^{(c)} = b^{(c)} + \sum_l \left( \alpha_l^{(c)} \sum_{i,j,k} A_{i,j,k}^{(l)} \right) \quad (8)$$

which can be reorganized as:

$$y^{(c)} = b^{(c)} + \sum_{i,j,k} \sum_l \alpha_l^{(c)} A_{i,j,k}^{(l)} \quad (9)$$

Finally, we use Equation 4 to rewrite Equation 9 as:

$$y^{(c)} = b^{(c)} + \sum_{i,j,k} (L_A^{(c)})_{i,j,k} \quad (10)$$

which is exactly the formula for sum-to-score property that we have declared.

**Making Approximations**. Base on the observations from our classification models for our task, we have $b^{(c)}$ negligible compared to $y^{(c)}$, we approximately have:

$$y^{(c)} \approx \sum_{i,j,k} (L_A^{(c)})_{i,j,k} \quad (11)$$

For CNN models which contain multiple fully-connected layers, there is some non-linearity, thus the gradients of feature maps are no longer constant. Luckily enough, our experiment supports that sum-to-score property works well for our CNN models. However, there is yet no evidence that this approximation always works within a certain range of error.

**Why Respond-CAM tends to have a better sum-to-score property than Grad-CAM**. Actually, Grad-CAM meets sum-to-score property under this assumption: for any given feature map $l$, the following equation meets:

$$\frac{1}{Z} \sum_{i,j,k} A_{i,j,k}^{(l)} \frac{\partial y^{(c)}}{\partial A_{i,j,k}^{(l)}} \approx \left( \frac{1}{Z} \sum_{i,j,k} A_{i,j,k}^{(l)} \right) \left( \frac{1}{Z} \sum_{i,j,k} \frac{\partial y^{(c)}}{\partial A_{i,j,k}^{(l)}} \right) \quad (12)$$

where $Z$ is the number of voxels in a feature map. In such case, the Equation 3 approximates:

$$\alpha_l^{(c)} \approx \frac{1}{Z} \sum_{i,j,k} \frac{\partial y^{(c)}}{\partial A_{i,j,k}^{(l)}} \quad (13)$$

In that case, Grad-CAM yields the similar results with Respond-CAM. Otherwise, Grad-CAM hardly meets sum-to-score property, which was observed in our experiments.

## 2  The Architectures of CNNs

The architectures of CNN models adopted in our experiments, namely *CNN-1* and *CNN-2*, are presented in Figure 1. For Respond-CAM/Grad-CAM, the scalar $y^{(c)}$ is an entry of the vector $y$, while $A$ is the tensor containing 64 feature maps, as labeled in Figure 1.

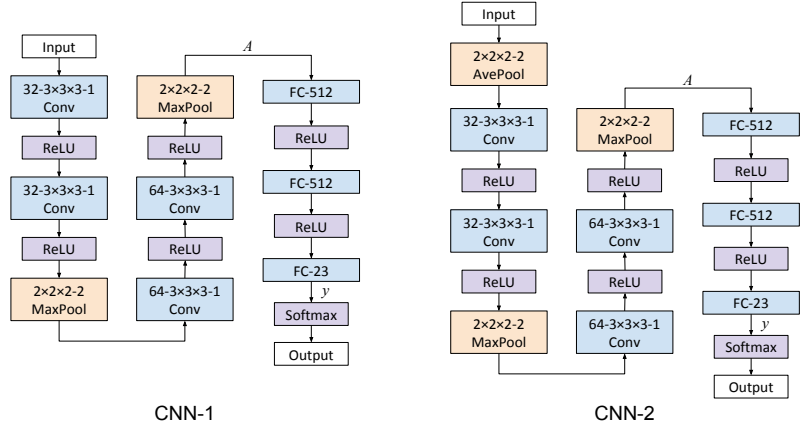

**Fig. 1.** Architecture of our CNN models. Each colored box represents one layer with its type and configuration. For example, "32-$3 \times 3 \times 3$-1 Conv" denotes a 3D convolutional layer with 32 filters, $3 \times 3 \times 3$ kernel size and stride 1. "$2 \times 2 \times 2$-2 MaxPool" or "$2 \times 2 \times 2$-2 AvePool" denote a 3D max pooling or average pooling layer over a $2 \times 2 \times 2$ region with stride 2, respectively. "FC-512" denotes a fully connected layer with an output vector of length 512. "ReLU" and "Softmax" denote different types of activation layers.

## 3 Respond-CAM on Natural Images

Our visualization approach may also be applied to various natural image computer vision tasks. The comparison results between Grad-CAM and our Respond-CAM is shown in Figure 2.

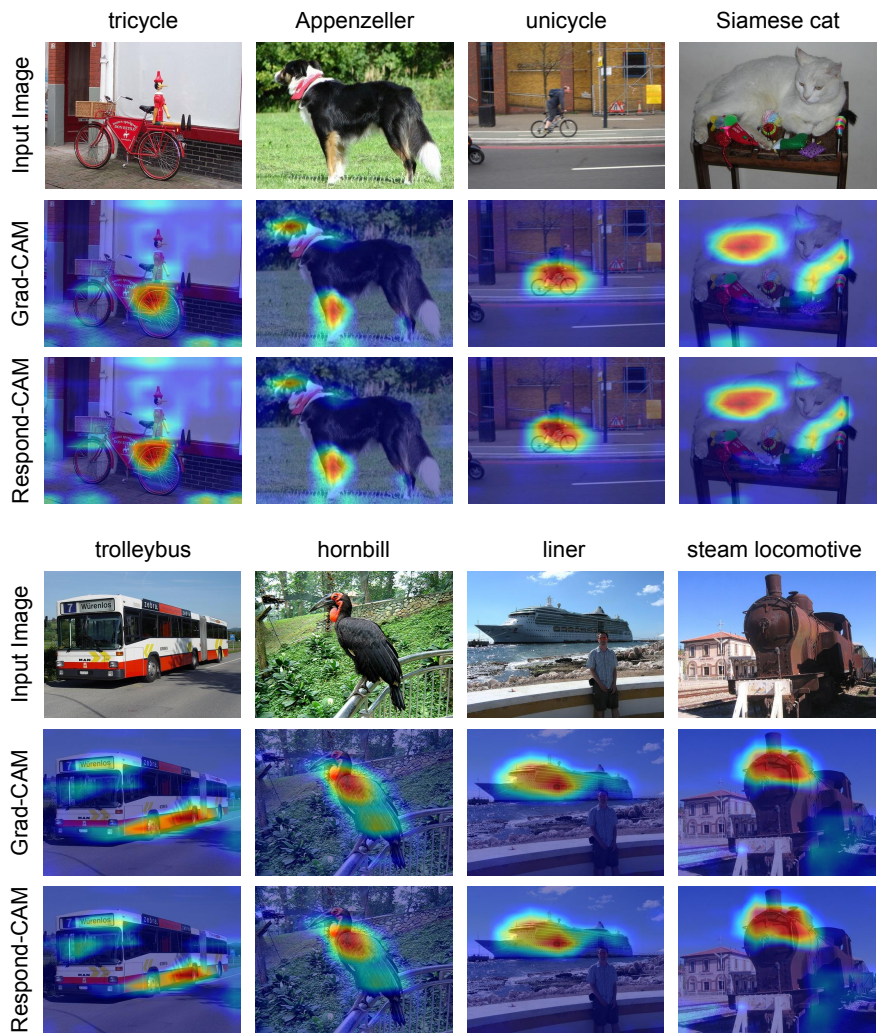

**Fig. 2.** Comparison of Grad-CAM and Respond-CAM on natural images. These images were selected from PASCAL 2007 test set. The VGG16 network with pre-trained parameters on ImageNet was used.